%% file: main.tex
\documentclass[10pt,twocolumn,letterpaper]{article}
\usepackage{cvpr}

\usepackage{dsfont}
\usepackage{times}
\usepackage{amsmath,amssymb} % define this before the line numbering.
\usepackage{graphicx}
\usepackage{caption}
\usepackage{subcaption}
\usepackage{algorithm}
\usepackage[noend]{algpseudocode}
\usepackage{tablefootnote}

\usepackage[pagebackref=true,breaklinks=true,letterpaper=true,colorlinks,bookmarks=false]{hyperref}

\cvprfinalcopy % *** Uncomment this line for the final submission

\def\cvprPaperID{6289} % *** Enter the CVPR Paper ID here

\ifcvprfinal\fi
\begin{document}

\title{Zoom-In-to-Check: Boosting Video Interpolation  via \\ Instance-level Discrimination
}

% The \author macro works with any number of authors. There are two
% commands used to separate the names and addresses of multiple
% authors: \And and \AND.
%
% Using \And between authors leaves it to LaTeX to determine where to
% break the lines. Using \AND forces a line break at that point. So,
% if LaTeX puts 3 of 4 authors names on the first line, and the last
% on the second line, try using \AND instead of \And before the third
% author name.

% NOTE: authors will be visible only in the camera-ready (ie, when using the option 'final'). 
% 	For the initial submission the authors will be anonymized.

\author{Liangzhe Yuan$^{1}$\thanks{indicates equal contribution} \quad
Yibo Chen$^{1*}$ \quad
%\thanks{indicates equal contributions.}$,
Hantian Liu$^1$ \quad
Tao Kong$^{1,2}$ \quad
Jianbo Shi$^1$\\
$^1$University of Pennsylvania \qquad $^2$Tsinghua University\\
{\tt\small $\{$lzyuan,yibochen,lhantian,jshi$\}$@seas.upenn.edu},
{\tt\small taokongcn@gmail.com}}
% \author{
% % 	Shaohui Liu$^{1}$\thanks{indicates equal contribution} \quad  Xiao Zhang$^{2*}$ \quad Jianqiao Wangni\textsuperscript{2} \quad Jianbo Shi\textsuperscript{2} \\ 
% 	$^1$Tsinghua University \qquad $^2$University of Pennsylvania \\
% 	{\tt\small b1ueber2y@gmail.com, \{zhang7, wnjq,  jshi\}@seas.upenn.edu}
% 	}

\maketitle
%\thispagestyle{empty}
%===============================================================================

% footnotes
% {\let\thefootnote\relax\footnote{{* indicates equal contributions.}}}
{\let\thefootnote\relax\footnote{{Supplementary video: \url{https://youtu.be/q-_wIRq26DY}.}}}

\begin{abstract}
We propose a light-weight video frame interpolation algorithm.  
Our key innovation is an instance-level supervision that allows information to be learned from the high-resolution version of similar objects.  Our experiment shows that the proposed method can generate state-of-the-art results across different datasets, with fractional computation resources (time and memory) of competing methods. 

Given two image frames, a cascade network creates an intermediate frame with 1) a flow-warping module that computes coarse bi-directional optical flow and creates an interpolated image via flow-based warping, followed by 2) an image synthesis module to make fine-scale corrections.
In the learning stage, object detection proposals are generated on the interpolated image.
Lower resolution objects are zoomed into, and the learning algorithms using an adversarial loss trained on high-resolution objects to guide the system towards the instance-level refinement corrects details of object shape and boundaries.
%As all our proposed network modules are fully convolutional, Our system can be trained end-to-end.
\end{abstract}
\vspace{-10pt}

%===============================================================================
% \begin{figure}
% \centering
% \includegraphics[width=1\linewidth]{figs/title_img.png}
% \caption{Video frame interpolation results. The first row: overlayed input frames; the second row: interpolated image; the third row from left to right: zoomed-in interpolated image, flow estimation and corresponding blending mask.}
% \label{fig:title}
% \vspace{-10pt}
% \end{figure}
\input{tex/Introduction.tex}
\input{tex/RelatedWork.tex}
\input{tex/Method.tex}
\input{tex/Experiments.tex}
\section{Conclusions}
We demonstrate a lightweight video interpolation framework that can retain instance level object details. We use a flow estimation module to synthesize the intermediate frame followed by a light-weight image synthesis module to correct detailed shape errors.
The network is trained by a region based discriminator which utilizes high-resolution image patches to supervise low-resolution RoIs, constraining instances in images to look realistic. 
%Our proposed pipeline has a zoom-in ability which gives more focus on small objects and ambiguous motions. 
Due to the modularity, our proposed adversarial training strategy can be universally used as a training block to improve algorithm performance. 
In the future, we hope to improve the model design to compensate some drawbacks in our model, e.g. employing deformable convolutions to tackle large motions and complex deformations. 
We also want to further expand our work to video prediction task.

\section{Acknowledgement}
We gratefully appreciate support through Honda Research Institute Curious Minded Machine program.

\newpage
{\small
% \nocite{*}
\bibliographystyle{ieee}
%\bibliography{refs}
\bibliography{egbib}
}

\end{document}

%% file: tex/Introduction.tex
\section{Introduction}
\label{sec:intro} 
\begin{figure}[!t]
\centering
\includegraphics[width=1\linewidth]{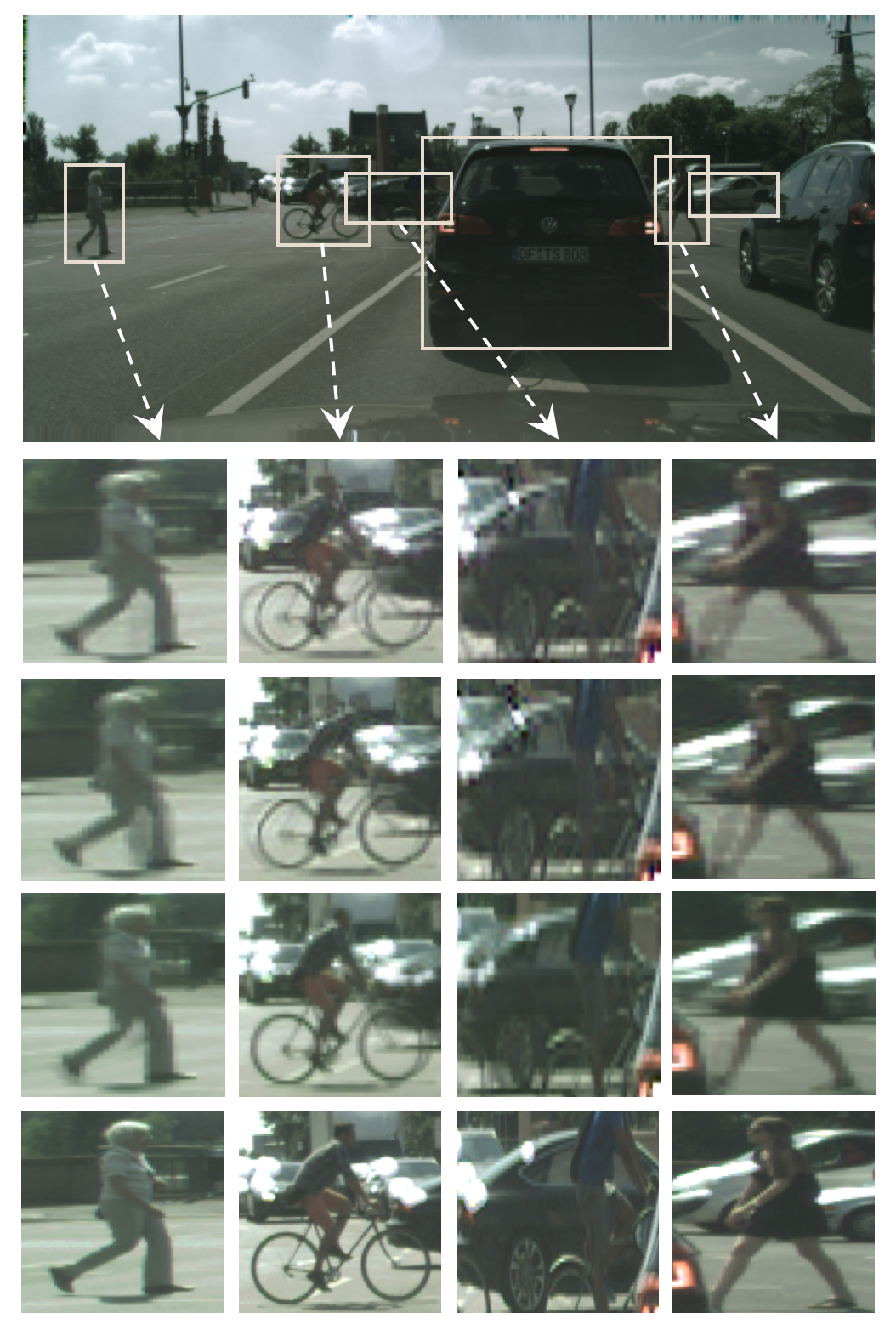}
\vspace{-20 pt}
\caption{
Object detection proposals allow region-of-interest (RoI) Zoom-In-to-Check.  An adversarial discriminator is trained using high-resolution objects across the entire video against the current interpolated image region. 
From top to bottom: synthesized image by image synthesis loss only; by whole image adversarial discriminator; by proposed instance-level discriminator; and the ground truth.
%Illustration of region proposals on image and the pooled region of interest, which are further sent into discriminator for instance level adversarial training. From top to bottom: synthesized image by Ours$_{roigan}$, pooled region from above, corresponding pooled region from Ours$_{gan}$ ,corresponding pooled region from Ours$_{baseline}$ and corresponding pooled regions from high resolution images used for training. Best view in color.
}
\label{fig:zoom}
\vspace{-20 pt}
\end{figure}
High fidelity video frame interpolation has usages in novel-view rendering, video compression, and frame rate conversion.
Existing methods focus on achieving overall high-quality interpolation averaged over all regions of images. The lack %lacking 
of explicitly object instances modeling became the bottleneck for algorithm's improvement.
% Existing methods tackle this problem using either motion estimation or image synthesis techniques, or the combination of both.

Flow-based image synthesis algorithms \cite{Baker2011,jiang2017superslomo,Niklaus_CVPR_2018,Reda2018sdcnet} generate realistic colors and patterns by explicitly copying pixels from given frames. 
For challenging scenes with occlusion, complex deformation or fast motion, flow-based interpolation suffers due to inaccuracy in optical flow estimation algorithms. 
To compensate for optical flow error,  \cite{Niklaus_CVPR_2018, Reda2018sdcnet} added an additional network to refine the interpolation results, at a cost of much higher computational cost. 

A kernel-based interpolation approach achieves the same per-pixel mapping goal without requiring a precise per-pixel flow estimation. The size of blending kernels in a such method directly restricts the motion that the network is able to capture.
To capture larger motion, big kernels ($51\times51$) are used in \cite{Niklaus_CVPR_2017}, which results in heavy memory and computation resource usage. 

%to learn a per-pixel mapping from input images to the target frame 
To hallucinate pixels on dis-occluded objects \cite{liang2017dualmotiongan} or sharpen motion blurred objects \cite{larsen2015autoencoding} often a Generative Adversarial Network (GAN) is used. 
However, such generative models are susceptible to mode collapse, resulting in over-fitting issues: when an object is blurry, it favors removing the object altogether. 

%Instead of generating pixel by copying from nearby frames,
We propose a lightweight video synthesis framework that takes advantage of a newly proposed instance-level adversarial training.  Our system consists of a two-stage interpolation network: a cascade design with a flow-based module followed by a kernel-based module. The design substantially alleviates the computational resource at the inference stage, as it requires neither large scale network needed to estimate accuracy optical flow, nor large size kernel needed to preserve clean boundary and capture large motion.

From our experiments, we found image-level supervision has a tendency to remove object details, particularly when the optical flow is fuzzy.  To alleviate this issue, we propose an instance-level discriminator to focus our system on the fine details of individual objects. 
However, if the `ground-truth' reference images also lack details due to lower resolution or motion blur, there is no sufficient feedback to the network on how to correct its mistakes.  Our key observation is that in the video we often have similar objects that appear at high-resolution with greater details.  This allows the algorithm to learn not just from the current reference frame, but also from semantically similar objects at higher-resolution.  

This design allows our network to leverage instance-level attention in learning and thus performs better in challenging scenes.  To the best of our knowledge, we are the first to present an instance-level adversarial learning framework that effectively exploits the network's capacity and achieves an accuracy-speed trade-off for video interpolation task. 
Using $\mathbf{78\%}$ computational time and 
$\mathbf{21\%}$ model parameters of SepConv \cite{Niklaus_ICCV_2017}, we achieve state-of-the-art interpolation quality. 

%% file: tex/RelatedWork.tex
\section{Related Work}
\label{sec:related_work}
Optical flow estimation is a basic building block for video frame interpolation \cite{jiang2017superslomo,Niklaus_CVPR_2018, wang2017light,liu2017voxelflow}.  
In fact, the image interpolation quality has been used to evaluate the accuracy of optical flow estimation \cite{Baker2011}. 
With rapidly improving quality of optical flow estimation, state-of-the-art optical flow methods \cite{dosovitskiy2015flownet,ilg2017flownet,sun2018pwc} can serve as a strong baseline for video interpolation.  
The drawbacks for flow-based video interpolation include 1) producing artifacts around object boundary due to lack of occlusion reasoning, 2) training optical flow estimators requires task-specific datasets, and 3) the overall algorithm is not end-to-end trainable.

One line of research focused on integrating optical flow into an end-to-end trainable video interpolation framework.  Liu \textit{et al.}\cite{liu2017voxelflow} developed a network to extract per-pixel 3D optical flow vector across space and time in the input video. The intermediate image is generated by trilinear interpolation across the input video volume. 
The method obtains high-quality results in frame interpolation and their unsupervised flow estimation results are comparable to the state-of-the-art.  However, \cite{liu2017voxelflow} tends to fail when the scene contains repetitive patterns.
The work by Jiang \textit{et al.} \cite{jiang2017superslomo} addressed the issue of occlusion by estimating bidirectional flow together with visibility mask, followed by a flow refinement network. 
Niklaus \textit{et al.} \cite{Niklaus_CVPR_2018} addressed the issue of inaccuracy of optical flow by retaining pixel-wise contextual information extracted from ResNet18 \cite{He2015resnet}, and employ a synthesis network with a GridNet \cite{fourure2017gridnet} architecture to generate the interpolated frame.

Moving away from optical flow based methods, \cite{meyer2018phasenet,Niklaus_ICCV_2017,Niklaus_CVPR_2017} eliminated the need of per-pixel explicit motion estimation.
Meyer \textit{et al.} \cite{meyer2018phasenet} propagate predicted phase information across oriented multi-scale pyramid levels to cope with large motions.
Niklaus \textit{et al.} \cite{Niklaus_CVPR_2017} estimate a spatially-adaptive convolution kernel for pixel synthesis for interoperation of two input frames.  
Although this method enables high-quality video frame interpolation, it is difficult to estimate all the kernels at once and the interpolation process is very memory intensive. 
In \cite{Niklaus_ICCV_2017}, the authors improve the efficiency by approximating a 2D kernel with a pair of 1D kernels. 
This work relieves the intensive memory requirement but the fundamental limitation still exists, where the capability of capturing large motion and flexibility in frame resolution is still limited by the kernel size, which is prohibitively expensive to increase.

A related but harder task is video frame extrapolation.  This task contains a similar challenge of motion estimation and object completion on dis-occluded regions. 
Earlier approaches use variational models that can represent the inherent uncertainty in prediction. 
Mathieu \textit{et al.} \cite{mathieu2015beyondmse} developed a multi-scale conditional GAN architecture to improve the prediction.  These methods suffer from blurriness and contain artifacts for large motion.
Vondrick \textit{et al.} \cite{vondrick2016generatingwithscendynamics} train a two-stream adversarial network that untangles foreground from background to predict into the future. 
Lee \textit{et al.}\cite{lee2018savp} propose a stochastic video prediction model based on VAE-GAN for object sythnesis and completion.
Several recent works seek to learn a transformation from past pixels to the future directly. 
\cite{vondrick2017generatingfuture} untangles the memory of the past from the prediction of the future by learning to predict sampling kernels. 
% \cite{jiang2017superslomo} can be extended to video extrapolation since it learns offset vectors for sampling. 
\cite{Reda2018sdcnet} combines flow-based and kernel-based approaches to learn a model to predict a motion vector and a kernel simultaneously for each pixel. 

%% file: tex/Method.tex
\section{Method}
\label{sec:method}
% We first introduce our forward video interpolation model, flow estimation module in Sec.\ref{sec:flow_mod} and image synthesis module in Sec.\ref{sec:synth_mod}. We discuss our region-of-interest (RoI) discriminator in Sec.\ref{sec:disc}, which is used to facilitate training our synthesis model. In Sec.\ref{sec:obj} we describe the overall losses used during training. Finally, we show our training details in Sec.\ref{sec:training}.

% \begin{figure}[!t]
% \centering
% \includegraphics[width=1\linewidth]{figs/final_roi_baseline.png}
% \caption{Illustration of region proposals on image and the pooled region of interest, which are further sent into discriminator for adversarial losses. From top to bottom:interpolated image by Ours$_{roigan}$, corresponding pooled region from Ours$_{gan}$, pooled regions from above image, corresponding pooled region from Ours$_{baseline}$. Best view in color.}
% \label{fig:zoom}
% \vspace{-10pt}
% \end{figure}
\input{tex/method/arch_fig.tex}
\input{tex/method/Architecture.tex}
\input{tex/method/ROIdiscriminator.tex}
\input{tex/method/Objectives.tex}
\input{tex/method/Training.tex}

%% file: tex/method/arch_fig.tex
\begin{figure}[!t]
\centering
\includegraphics[width=1\linewidth]{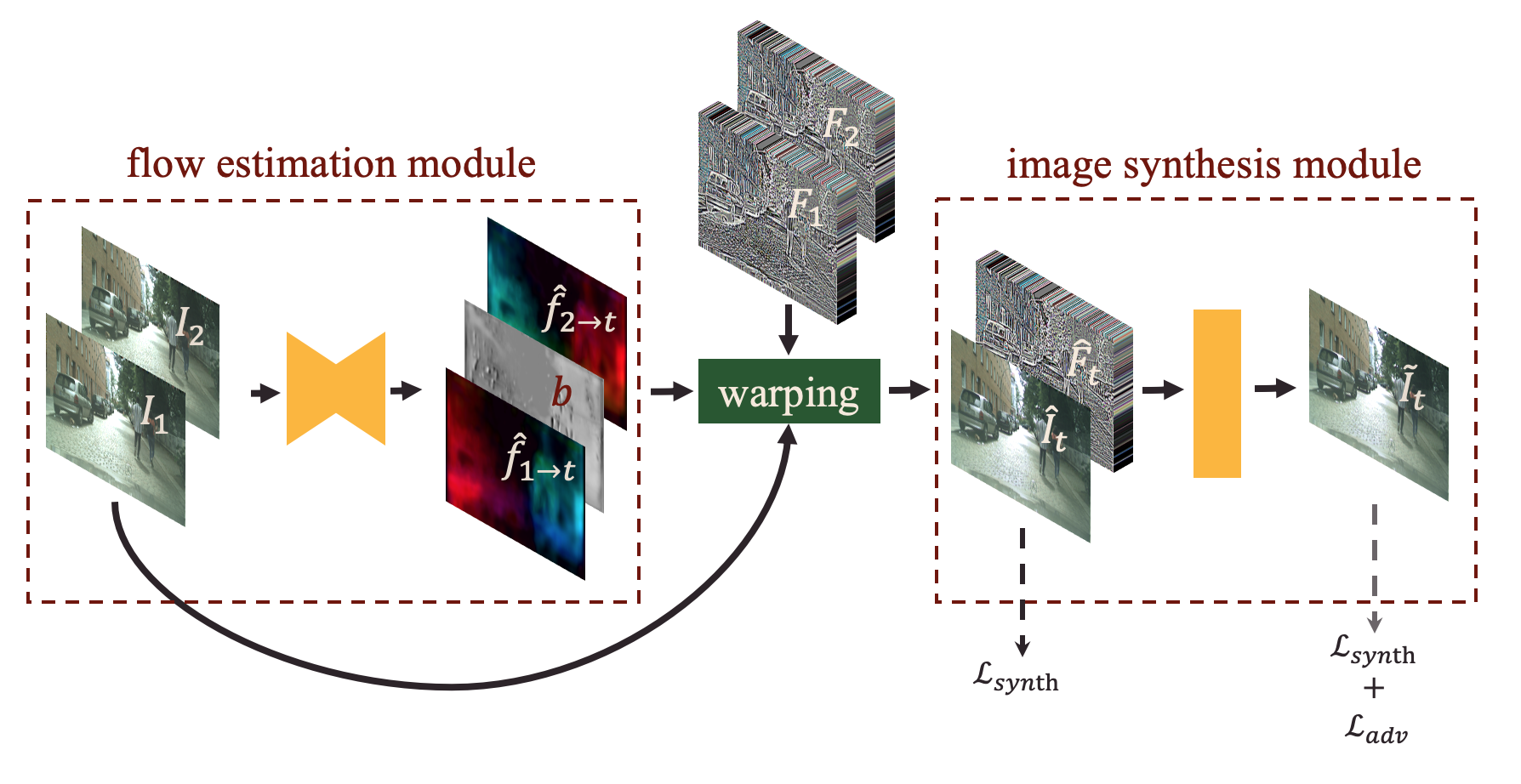}
\caption{
An overview of our model. 
The flow estimation module (left) takes two frames: $\mathbf{I}_1$ and $\mathbf{I}_2$ as input. 
It predicts the bidirectional optical flows $f_{1 \rightarrow t}$ and $f_{2 \rightarrow t}$ for coarse motion estimation, and a blending mask $b$ for occlusion reasoning. 
The image synthesis module (right) takes images $\mathbf{I}_1, \mathbf{I}_2$, corresponding features $\mathbf{F}_1$, $\mathbf{F}_2$, estimated optical flows $f_{1 \rightarrow t}$, $f_{2 \rightarrow t}$ and blending mask $b$ to synthesize target frame $\Tilde{\mathbf{I}}_t$. 
Instance-level adversarial discrimination is further added on $\Tilde{\mathbf{I}}_t$ to preserve sharper image details. 
% The adversarial learning structure is described in Fig.\ref{fig:roi-training}.
% The final region discrimination module examines specific regions of interest in $\Tilde{\mathbf{I}}_t$ and compares them against high resolution counterparts.
% %to guide generating more realistic image. 
% Only flow estimation module and image synthesis module are used during inference stage.
}
\label{fig:gen}
\vspace{-10pt}
\end{figure}

%% file: tex/method/Architecture.tex
\subsection{Coarse Optical Flow Estimation}
\label{sec:flow_mod}
To compensate large displacement motion, we first estimate coarse optical flow to generate an initial interpolated frame $\hat{\mathbf{I}}_t$ given two consecutive video frames $\mathbf{I}_1$ and $\mathbf{I}_2$.
% Given two consecutive video frames $\mathbf{I}_1$ and $\mathbf{I}_2$, our goal is to generate an intermediate frame $\mathbf{I}_t$ in between the two input frames. 
%Our generator network can be divided into two sub-modules as illustrated in Fig.\ref{fig:gen}, namely the flow estimation module and the image synthesis module. 
% Additionally, in order to tackle with common challenging scenarios for video interpolation tasks, such as fast motion, cluttered environment, severe occlusion,etc, we propose to use another network to discriminate the synthesis results. %? delete?
We use a U-Net like network to estimate bidirectional optical flows $f_{1 \to t}$ and $f_{2 \to t}$, which can be used to warp $\mathbf{I}_1$ and $\mathbf{I}_2$ respectively to %destined 
designated time $\hat{\mathbf{I}}_t$. 
In the meantime, our network also predicts a per-pixel weighting mask $b$ to blend two synthesized images into one. The blending mask $b$ here can be %understand 
seen as a confidence mask and it's designed to deal with occlusion.
Inspired by \cite{Niklaus_CVPR_2018}, we employ a pre-trained feature extractor to extract high level features from both $\mathbf{I}_1$ and $\mathbf{I}_2$, denoted as $\mathbf{F}_1$ and $\mathbf{F}_2$ respectively. 
Note that empirically the flow-based methods present a satisfying performance on most of the regions but %it has difficulty dealing with
often fail to cope with fine-grain details and complex motions. 
Thus our flow estimation module only serves as an initial step for video interpolation task.

\subsection{Image Synthesis Module}
\label{sec:synth_mod}
We perform both pixel-level and semantic feature-level warping as shown in Fig.\ref{fig:gen}. 
In detail, we feed images $\mathbf{I}_1$, $\mathbf{I}_2$, corresponding deep feature maps $\mathbf{F}_1$, $\mathbf{F}_2$, flows $f_{2 \to t}$, $f_{1 \to t}$ and mask $b$ into later module for further refinement. 
In the image synthesis module, we use estimated bi-directional flow $f_{2 \to t}$, $f_{1 \to t}$ and blending mask $b$ to warp both images and features into time $t$ by bi-linear interpolation \cite{jaderberg2015spatial}.
\begin{align}
    \hat{\mathbf{I}}_{t} = b \odot g(\mathbf{I}_1, f_{1 \to t}) + (1-b) \odot g(\mathbf{I}_2, f_{2 \to t})\\
    \hat{\mathbf{F}}_{t} = b \odot g(\mathbf{F}_1, f_{1 \to t}) + (1-b) \odot g(\mathbf{F}_2, f_{2 \to t}) 
\end{align}
where $g(\mathbf{I}, f)$ is the bi-linear warping function that takes a warping map $f$ to warp a tensor $\mathbf{I}$ to $\hat{\mathbf{I}}$
% \begin{align}
%     g(\mathbf{I}, f) =& \sum_{i=0}^{H \times W} \sum_{j=0}^{H \times W} I_j \mathbf{max}(0, |f^u_i - u_j|)  \mathbf{max}(0, |f^v_i - v_j|) \nonumber\\
%     &\forall I_j \in \mathbf{I}, f_i \in f
% \end{align}
and $\odot$ is an element-wise multiplication operator.
Then we concatenate warped features $\hat{\mathbf{F}}_t$ and image $\hat{\mathbf{I}}_t$ and feed it into the image synthesis layers. Different from  \cite{Niklaus_CVPR_2018}, in which the author used a giant GridNet \cite{fourure2017gridnet} to refine the image, we simply use three convolutional layers with kernel size 9 to approximate a large receptive field. We will show that this approximation is enough to get good performance with our proposed instance-level adversarial loss.

%one deformable convolution layer to produce final image synthesis $\Tilde{I}_t$. %As we know that 
%Deformable convolution \cite{dai17deformableconv} predicts an offset map for each kernel at each pixel, which implicitly learns a local re-sampling map. It is empirically sufficient to recover the blurriness and rectify the distortion due to the inferior former estimation of optical flow. %Formally, 
% The pixel synthesis using deformable convolution is computed as
% \begin{align}
%     \Tilde{I}_t(p) = \sum_{p} K(q)\hat{I}_t(p + q + \Delta \mathbf{p}(p, q))
% \end{align}
% where $K$ is the convolutional kernel, $p$ is the enumerator on image, $q$ is the enumerator on convolutional kernel and $\Delta \mathbf{p}(p,q)$ is the estimated offset map in deformable convolution. 

%% file: tex/method/ROIdiscriminator.tex
\subsection{Instance-level Discriminator}
\label{sec:disc}
\input{tex/exp/roi-training.tex}
Flow-guided warped image $\hat{\mathbf{I}}_t$ generated from the two previous stages has two problems:
(a) as the optical flow is trained on the whole image, it often results in twisted and blurry boundaries, as shown in Fig.\ref{fig:zoom}; 
% b) the bi-linear interpolation result is less realistic compared with the original target image $\mathbf{I}_t$. 
% b) without instance awareness of the model, the interpolation results are less realistic and the entities in the images are tend to be removed to minimize the training objectives. 
b) optical flow estimation fails to dis-occlude objects in the images, which are common cases the interpolation algorithm needs to deal with.
% To address the issues, we propose to focus the learning algorithm on instances with various size and deformation in the images using adversarial learning \cite{Goodfellow2014GAN}.
To address the issues, we use the adversarial learning \cite{Goodfellow2014GAN} to empower the model on synthesizing instances and recovering structural patterns.
In the experiments, we explore two algorithm variations for video interpolation: 
(a) directly discriminating on the whole image, and (b) zoom-in on object instance area, as shown in Fig.\ref{fig:roi-training}. 
% a typical practice is computing adversarial losses on the whole image, which leads to image-level adversarial training. 

Direct adversarial learning on the whole image makes the generated $\tilde{\mathbf{I}}_t$ looks more realistic compared with the real image $\mathbf{I}_t$.
However, since the majority of the image is usually the background, the image-level supervision provides a uniform gradient across the whole image, such that the semantic details are ignored and the optimization of the foreground is diminished.
% since the majority of the image is usually the background, the adversarial loss of Eq.\ref{eq:gan} will be biased which hurts the optimization of foreground areas. 
% One example is shown in Fig.\ref{fig:zoom} row 2 and 3.

The instance-centered learning forces the model to pay more attention to instances, especially on small-scale objects. 
Given an image $\mathbf{I}_t$, we use region proposal method \cite{he2017maskrcnn} to generate several regions of interest (RoI).
If we have access to the high-resolution images during training, we crop the corresponding RoIs from high-resolution images and use them to guide the synthesis of low-resolution results. 
We perform \textit{RoIAlign} as described in \cite{he2017maskrcnn} to pool the RoIs from $\mathbf{I}_t$ and $\hat{\mathbf{I}}_t$ into patches with fixed size of $h \times w$.
The \textit{RoIAlign} can achieve two effects: a) through bi-linear interpolation, the gradient can be backpropagated to the exact pixel location and previous modules, thus the total network can be updated end-to-end; 
b) reshaping operation naturally realizes zoom-in effect, balancing network's focus on close and far away, large and small objects. 
The reshaped RoIs of different objects are illustrated in Fig.\ref{fig:roi-training}.

There are two ways to choose how many RoIs per image used for training: we can either choose a fixed number of RoIs with highest response from region proposal network, or use RoIs whose score are above a certain threshold of non-maximum suppression during region proposing.
In our experiments, we found the final interpolation quality is not sensitive to the number of RoIs per image used for training. 
Using 10-30 region proposals per image during training leads to $\pm0.002$ \textit{STD} of SSIM and $\pm0.08$ \textit{STD} of IE/PSNR in testing.
%We show this in Table \ref{tab:num_bbox}. 
% In this paper, 
We empirically choose 16 RoIs per image in the training stage.

% Generated RoI numbers are inconsistent among different images, which makes it hard to construct data batches fed into the network. 
% To solve the issue, a fixed region proposal number is assigned to each image, and empty proposal spaces left in each image are filled in with zeros. 
% An extra number recording valid RoI numbers in each image is fed into the network as well, in order to guarantee that only valid RoIs are processed in later discrimination process. 
A discriminator with spectral normalization \cite{miyato2018spectral} is employed to examine only on the specific RoIs instead of on the whole image. 
The details of adversarial loss $\mathcal{L}_{adv}$ are described in the next section.

%% file: tex/exp/roi-training.tex
\begin{figure}[!t]
 \centering
      {\includegraphics[width=1\linewidth]{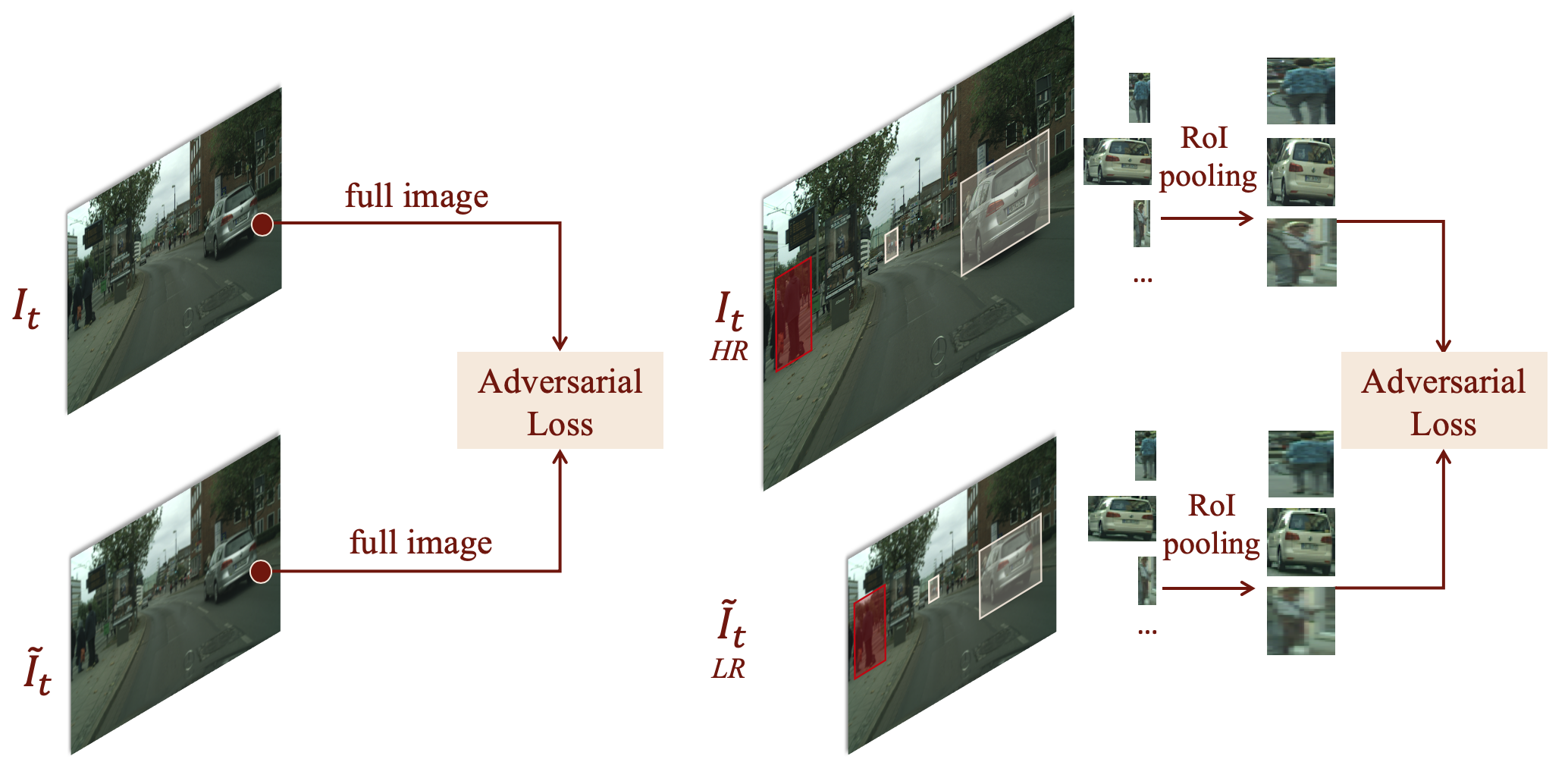}}
  \centering
  \caption{Image level adversarial learning v.s.  proposed instance level adversarial learning. We crop RoIs from high resolution images and resize them into constant size patches, which are used to train our low resolution images. This forces the system to focus on refining details and boundaries of instances.}
  \label{fig:roi-training}
  \vspace{-10pt}
\end{figure}

%% file: tex/method/Objectives.tex
\subsection{Training Objectives}
\label{sec:obj}
\input{tex/exp/bar_runtime.tex}

We use two losses to train the network: a global interpolation loss $\mathcal{L}_{int}$ and an instance adversarial loss $\mathcal{L}_{adv}$. 
%This is reflected in our loss function that consists of two terms: a synthesis loss $\mathcal{L}_{synth}$ and an adversarial loss $\mathcal{L}_{adv}$. The overall objective we use to train our system is:
\begin{align}
    \mathcal{L} = \mathcal{L}_{int} + \mathcal{L}_{adv}
\end{align}

\textbf{Interpolation Loss.}
For the global interpolation loss, we first minimize the robust $\ell 1$ norm \cite{sun2014quantitative} on the per-pixel color difference, which is used in recent self-supervised optical flow estimation work \cite{zhu2018ev}. 
We further constrain the first-order gradient difference between interpolated image and the groundtruth to be consistent, which further improves the reconstruction quality \cite{mathieu2015beyondmse}. 
The above photometric losses are computed as
\begin{align}
    \mathcal{L}_{ph} =& \rho(\Tilde{\mathbf{I}} - \mathbf{I}_{gt}) \nonumber\\ 
    +& \rho(\frac{\partial \Tilde{\mathbf{I}}}{\partial x} - \frac{\partial \mathbf{I}_{gt}}{\partial x})
    + \rho(\frac{\partial \Tilde{\mathbf{I}}}{\partial y} - \frac{\partial \mathbf{I}_{gt}}{\partial y})\\
    \rho(x) =& (x^2 + \epsilon^2)^\alpha
\end{align}
where $\rho(\cdot)$ is the robust $\ell 1$ norm also known as Charbonnier norm.

The second term of the interpolation loss is perceptual loss \cite{johnson2016perceptual}. It quantifies the network higher-level feature reconstruction quality and thus makes more visually plausible image interpolation results. 
Our experiments show that the perceptual loss enables the network to learn to reconstruct crispy image boundary. 
The perceptual loss is defined as
\begin{align}
    \mathcal{L}_{pe} = |\Phi(\Tilde{\mathbf{I}}) - \Phi(\mathbf{I}_{gt})|_1
\end{align}
in which the $\Phi(\cdot)$ is the feature extraction function and in our work, we use the latent features from VGG-16 \cite{vgg}.
We apply photometric loss and perceptual loss on both the initial interpolated image $\hat{\mathbf{I}}$ and the synthesized image $\Tilde{\mathbf{I}}$.
% to stabilize flow estimation network training. 
We also constrain the first-order gradient of bi-directional optical flow $f_{1 \to t}, f_{2 \to t}$ and the corresponding blending mask $b$ to be locally smooth, resulting in smoothness loss $\mathcal{L}_{s}$.

% To constrain the predicted bi-directional optical flow $f_{1 \to t}, f_{2 \to t}$ and the corresponding blending mask $b$ to be locally smooth, we introduce a smoothness loss $\mathcal{L}_{s}$
% \begin{align}
%     \mathcal{L}_{s} &= \sum_{i = 1,2} \rho(\frac{\partial f_{i \to t}}{\partial x})
%                             + \rho(\frac{\partial f_{i \to t}}{\partial y}) \nonumber\\
%                             &+ \rho(\frac{\partial b}{\partial x})+ \rho(\frac{\partial b}{\partial y})
% \end{align}

%the $\lambda $ controls different smoothness loss weights on optical estimation and blending mask estimation.
The above loss functions applied on full images mainly guide our network for the coarse level interpolation and we group them as the interpolation loss,
\begin{align}
    \mathcal{L}_{synth} = \lambda_0 \mathcal{L}_{ph} + \lambda_1 \mathcal{L}_{pe} + \lambda_2 \mathcal{L}_{s}
\end{align}

\textbf{Adversarial Loss.} In order to deal with complex scenarios and enlarge model capacity, we utilize another network $D(\cdot)$ to discriminate synthesized images. 
The adversarial loss consists of two parts, namely the generator loss and the discriminator loss.
% As illustrated in Fig.\ref{fig:roi-training}, a typical practice is computing adversarial losses on the whole image, which leads to image-level adversarial training. 
% The image-level supervision provides a uniform gradient across the whole image, such that semantic details and fine-grained differences are ignored. 
% In contrast, we propose an instance-level adversarial learning. 
% For each synthesized and ground truth image pair $(\Tilde{\mathbf{I}}, \mathbf{I}_{gt})$, we use region proposal method to propose $N$ region proposals and resize them into $h \times w$ patches, resulting in %$(\Tilde{P}_i, P_i), i = 1, \cdots, N$ ROI pairs. 
% $N$ RoI pairs.
% As shown in Fig. \ref{fig:roi-training}, if we have access to the high-resolution images during training, we crop the corresponding RoIs from high-resolution images and use them to guide the synthesis of low-resolution results. 
Let the $(\Tilde{\mathbf{P}}_i, \mathbf{P}_i)$ refer to a pair of synthesized and groundtruth RoIs, where $i = 1, \cdots, N$.
The discriminator will examine each one of them and the adversarial losses are formulated as:
\begin{align}
\label{eq:gan}
    \mathcal{L}_{d} =& \frac{1}{N} \sum_{i = 1}^N  \mathbb{E}[\mathbf{min}(0, -1-D(\Tilde{P}_i))] \nonumber
                       \\+&
        \mathbb{E}[\mathbf{min}(0, -1+D(P_i))] \nonumber\\
    \mathcal{L}_{g} =&  -\frac{1}{N} \sum_{i = 1}^N \mathbb{E}[D(\Tilde{P}_i)];
    \mathcal{L}_{adv} = \lambda_3 \mathcal{L}_{d} + \lambda_4 \mathcal{L}_{g}
\end{align}

%% file: tex/exp/bar_runtime.tex
\begin{figure}[!t]
 \centering
      {\includegraphics[width=1\linewidth]{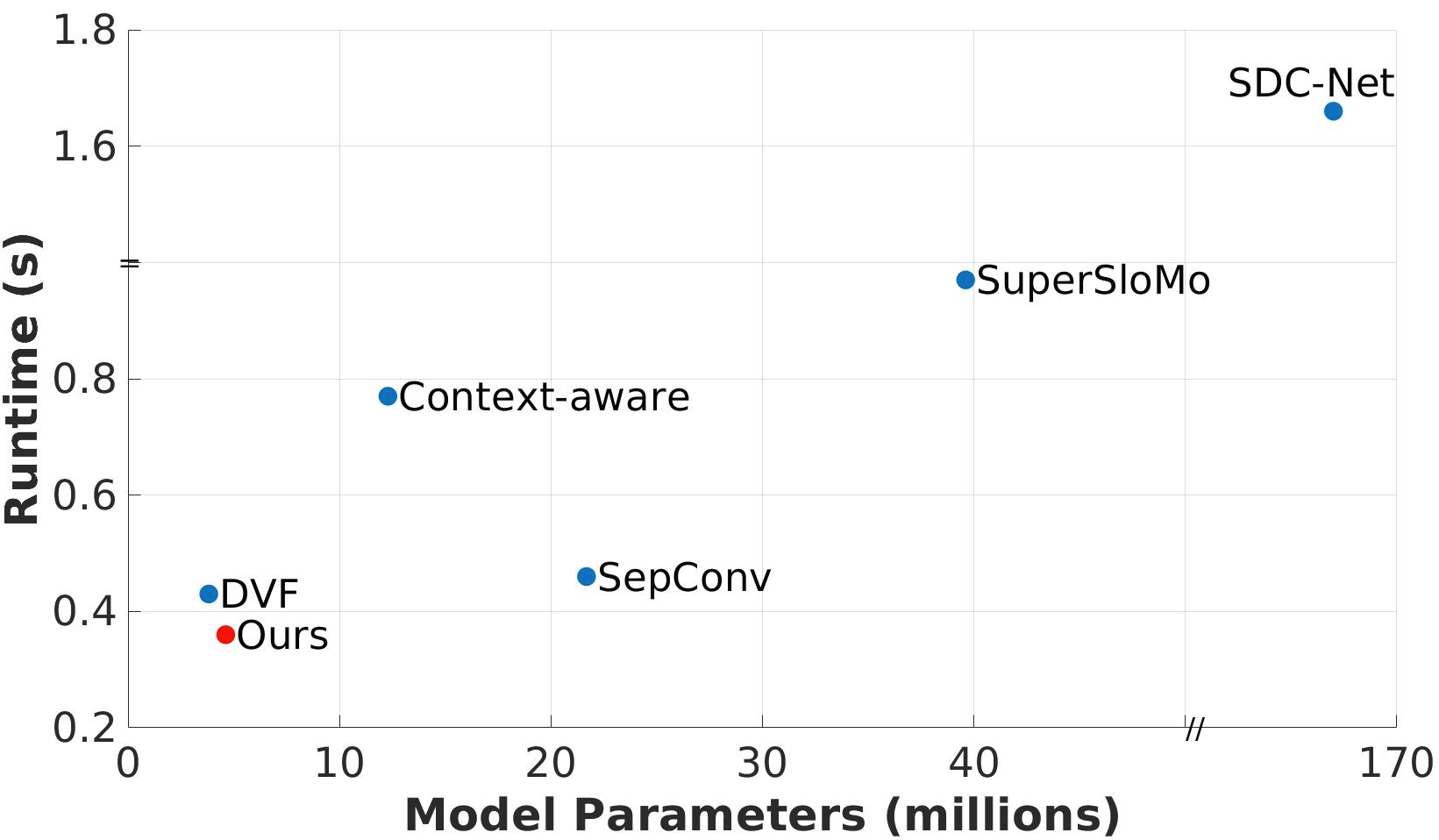}}
  \centering
  \caption{Trained with the proposed instance-level adversarial loss, our model generates the best results with minimal number of parameters and least run-time on interpolating FHD resolution images, comparing to other methods.}
  \label{fig:runtime}
  \vspace{-15pt}
\end{figure}

%% file: tex/method/Training.tex
\subsection{Training Details}
\label{sec:training}
\input{tex/exp/fig_trimap.tex}
\input{tex/exp/compare_trimap.tex}
\input{tex/exp/fig_ssimcomp.tex}
The network is trained on a mixer of UCF101 \cite{UCF101} and CityScapes \cite{Cordts2016Cityscapes} dataset.
We randomly pick four triplets in every video clip of UCF101 and one triplet in every sequence of CityScapes training set, which gives us around 26k triplets in all.
In practice, as our proposed training pipeline is self-contained and does not need labels, any collection of video clips are sufficient to train our network.
We keep UCF101 original image size and downsample CityScapes images to $256 \times 512$. % for training inputs. 
Note that we use the high-resolution version of images in CityScapes dataset to supervise adversarial learning. 
Forming high-resolution and low-resolution training pair is the key to our learning algorithm. 
During training, we randomly crop a $256 \times 256$ region of triplets as input.
We also randomly flip images for data augmentation. 
The size of output from \textit{RoIAlign} is set to be $64 \times 64$.
An Adam optimizer \cite{adam2014} with $\beta_1 = 0.9$ and $\beta_2 = 0.999$ is used with initial learning rate 1e-4, which is decayed exponentially by a factor of 0.1 for every 10 epochs and clipped at 1e-8 during training. 
Also, we added decayed random noise to `real' images and scheduled to train the discriminator more to smooth the adversarial learning.
%We place image synthesis loss on both initial interpolated image $\hat{I}_t$ and final synthesis image $\Tilde{I}_t$ and apply adversarial loss only on image $\Tilde{I}_t$. 
The weights for different losses are set as $(\lambda_0, \lambda_1, \lambda_2, \lambda_3, \lambda_4) = (1, 1, 0.01, 0.1, 0.01)$. 
% We trained our model with batch size 2 on a 11GB NVIDIA  GeForce 1080 Ti GPU for around 16 hours.

%% file: tex/exp/fig_trimap.tex
\begin{figure}[!t]
\centering
\vspace{-10pt}
\includegraphics[width=\columnwidth, height=3.2cm]{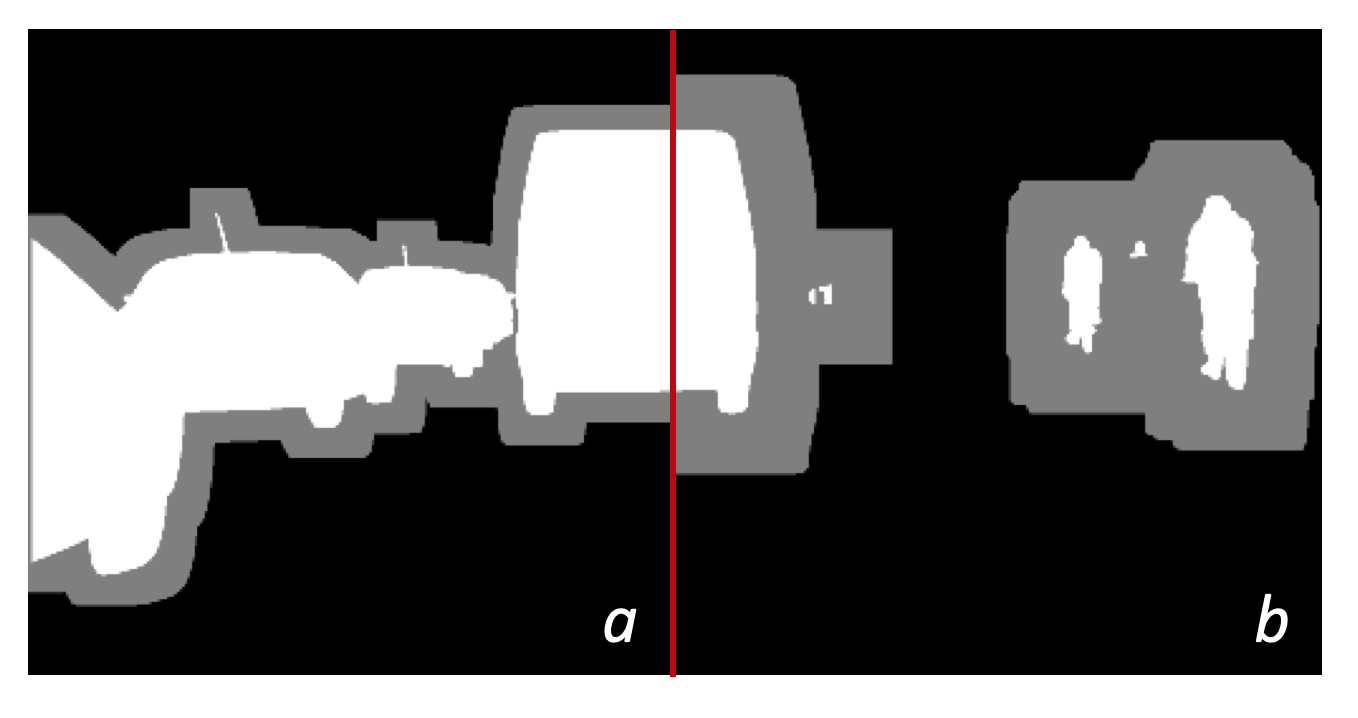}
\vspace{-15pt}
\caption{
% Left: 
Example of a trimap mask using groundtruth segmentation (white) with 12(a)/24(b) pixels dilation (gray). Evaluation is done on gray and white area. 
% Right: Ours$_{\text{roigan}}$ preserves more perceptual structures than SepConv.
}
\label{fig:example_trimap}
\vspace{-10pt}
\end{figure}

%% file: tex/exp/compare_trimap.tex
\begin{figure}[!t]
\centering
% \vspace{-5pt}
% \renewcommand{\thefigure}{1*}
\includegraphics[width=\columnwidth, height=4.6cm]{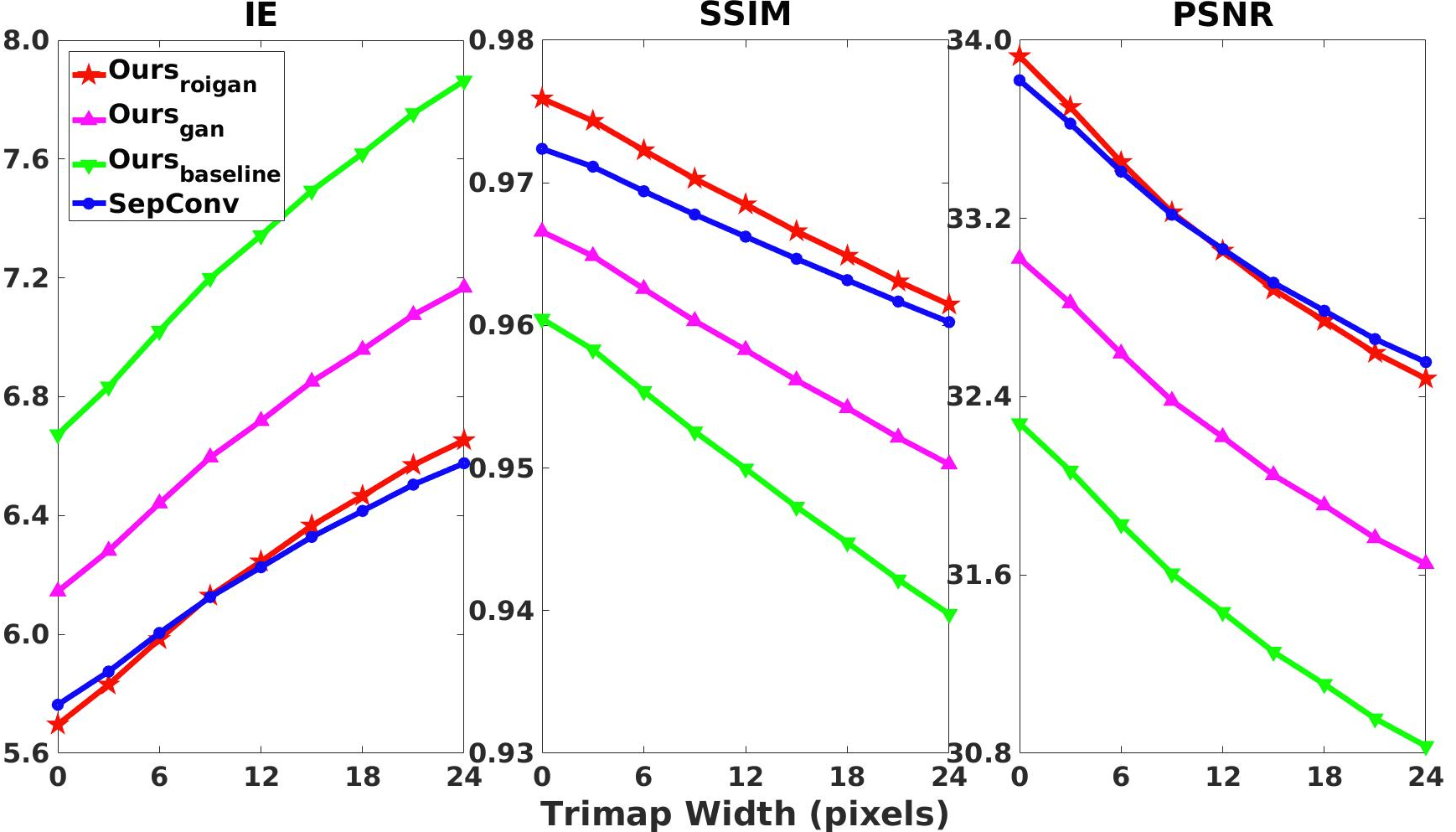}
%\vspace{-3pt}
\caption{Evaluation metrics on trimaps with various widths on CityScapes dataset. DVF \cite{liu2017voxelflow} is excluded in the plot for better visualization.}
\label{fig:metrics_trimap}
\vspace{-12pt}
\end{figure}

%% file: tex/exp/fig_ssimcomp.tex
\begin{figure}[!t]
\centering
% \vspace{-10pt}
% \renewcommand{\thefigure}{2*}
\includegraphics[width=\columnwidth, height=3cm]{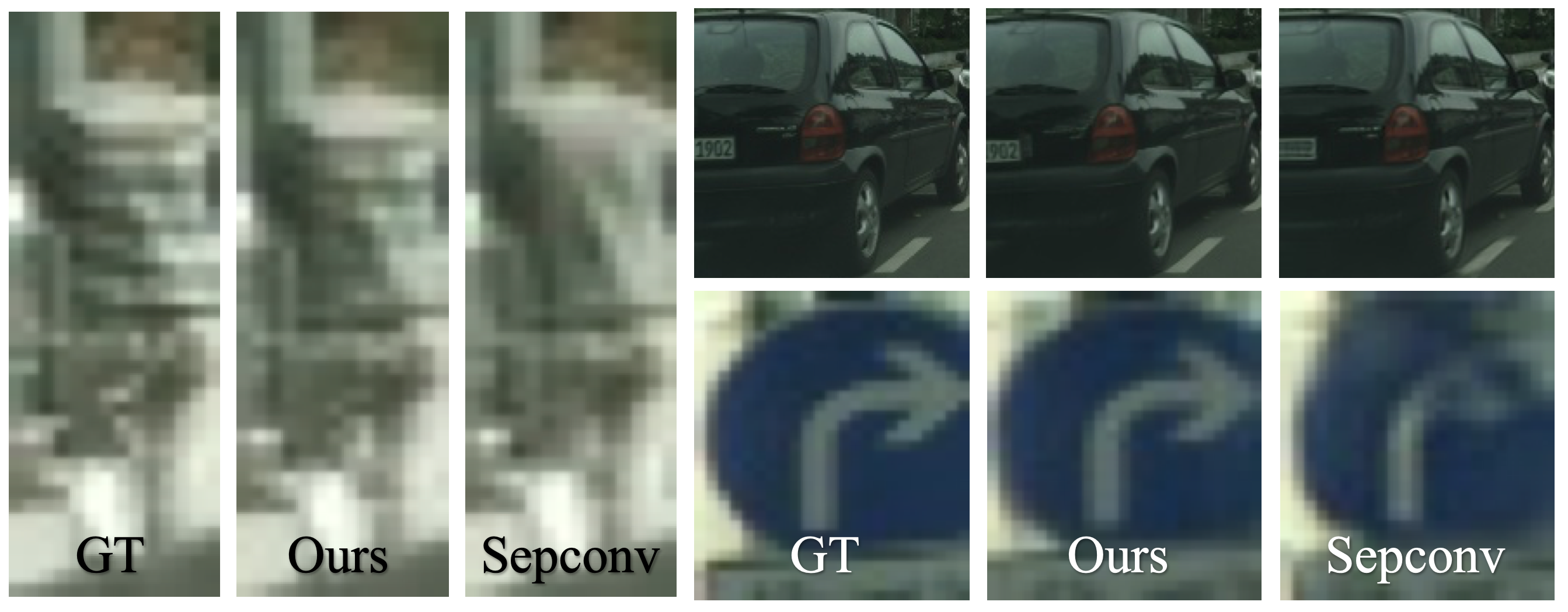}
\caption{
% SSIM measures structural and perceptual similarity.
Higher SSIM score from Ours$_{roigan}$ indicates we preserve more perceptual structures than SepConv \cite{Niklaus_ICCV_2017}.}
\label{fig:example_ssim}
\vspace{-15pt}
\end{figure}

%% file: tex/Experiments.tex
\section{Experiments}
\label{sec:experiments}
\input{tex/exp/table_merge_v.tex}
% \input{tex/exp/bar_runtime.tex}

% \begin{figure}[!t]
%  \centering
%       {\includegraphics[width=1\linewidth]{figs/Runtime_barchart.png}}
%   \centering
%   \caption{Runtime on $1920 * 1080$ image comparison between Ours$_{roigan}$, Context-aware Synthesis, SDC-Net}
%   \label{fig:runtime}
% \end{figure}

\input{tex/exp/compare_figs.tex}
\input{tex/exp/compare-ucf.tex}

To evaluate our method, we quantitatively and qualitatively compare it with several state-of-art video frame interpolation methods.
Namely, Deep Voxal Flow (DVF) \cite{liu2017voxelflow} is a flow warping method for video interpolation;
Seperable adaptive Convolutions (SepConv) \cite{Niklaus_ICCV_2017} is a kernel based method with the adaptive separable convolutions;
SuperSloMo \cite{jiang2017superslomo} uses a cascaded optical flow estimator to interpolate video frames. 
We refer Ours$_{roigan}$ as our network trained with proposed instance-level adversarial loss, Ours$_{gan}$ as the model trained with the adversarial loss on overall image, and Ours$_{baseline}$ as the model trained without any adversarial loss.

We compare the algorithm results on two different datasets, CityScapes \cite{Cordts2016Cityscapes} and UCF101 \cite{UCF101}. 
CityScapes contains different objects, e.g. cars, people, traffic lights, etc., with various size and distance, which is good to differentiate algorithms' interpolation abilities on small objects and partial occlusions.
UCF101 contains people activities, e.g. boating, making-up, boxing, etc., which is good to show results on fast motion and complex deformation. 
% In ablation experiments, we compare several variations of our proposed approach, basic flow warping structure, adversarial learning on the entire image structure, and instance-level adversarial learning network structure.  

\input{tex/exp/Ablation.tex}
\input{tex/exp/Quantitative.tex}
\input{tex/exp/Qualitative.tex}
\input{tex/exp/Discussion.tex}

%% file: tex/exp/table_merge_v.tex
\begin{table}[t!]
\centering
 \begin{tabular}{cccc} 
 \hline\\[-1ex]
UCF101 \cite{UCF101} & IE & SSIM & PSNR \\ [0.5ex] 
%   \cline{2-4} \\[-1.5ex] 
%   & \multicolumn{3}{c}{UCF101 \cite{UCF101}}\\ [0.5ex] 
 \hline \\[-1ex] 
DVF \cite{liu2017voxelflow} & 11.54 & 0.869 & 29.70\\ 
SepConv \cite{Niklaus_ICCV_2017} & 11.28 & 0.875 & 30.29  \\
SuperSloMo \cite{jiang2017superslomo}& \textbf{10.87} & \textbf{0.885} & \textbf{30.48}  \\
Ours$_{baseline}$ & 11.23 & 0.876 & 30.08    \\ % 0.876 not for sure 
Ours$_{gan}$  & 11.66 & 0.870 & 29.85 \\ 
Ours$_{roigan}$ & 10.92 & 0.882 & 30.23 \\ 
\hline\\[-1ex] 
%  & \multicolumn{3}{c}{CityScapes \cite{Cordts2016Cityscapes}} \\ [0.5ex]
CityScapes \cite{Cordts2016Cityscapes} & IE & SSIM & PSNR \\ [0.5ex] 
%  \cline{2-4} \\[-1.5ex] 
 \hline\\[-1ex]
 DVF \cite{liu2017voxelflow} & 17.49 & 0.722 & 23.88 \\
 SepConv \cite{Niklaus_ICCV_2017} & \textbf{7.85} & 0.923 &\textbf{30.92} \\
 SuperSloMo \cite{jiang2017superslomo} & $-$ & $-$ & $-$\\
 Ours$_{baseline}$ & 9.38  &0.890  &29.31\\
 Ours$_{gan}$ & 9.04 & 0.902 & 29.93\\
 Ours$_{roigan}$ & 8.03 & \textbf{0.925} & 30.77  \\
  \hline
\end{tabular}
\caption{Quantitative evaluation of different methods on CityScapes\protect\footnote[1]{} and UCF101\protect\footnote[2]{}, including Interpolation Error (IE) \cite{Baker2011}, Peak-Signal-To-Noise (PSNR), and Structural-Similarity-Image-Metric (SSIM). Lower IE and higher SSIM and PSNR indicate better quality. 
}
\vspace{-15pt}
\label{tab:merge}
\end{table}

\footnotetext[1]{SuperSloMo \cite{jiang2017superslomo} is not open-sourced so we don't have their results on CityScapes dataset.}
\footnotetext[2]{We re-run the evaluation on the synthesis images provided by \cite{jiang2017superslomo}.}

%% file: tex/exp/compare_figs.tex
\begin{figure*}[t!]
\centering
\setlength\tabcolsep{1.5pt} % default value: 6pt
\begin{tabular}{cccccccc}
 \multicolumn{2}{c}{\includegraphics[width=0.25\linewidth]{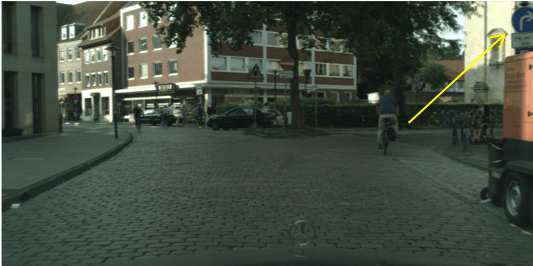}} &
  \multicolumn{2}{c}{\includegraphics[width=0.25\linewidth]{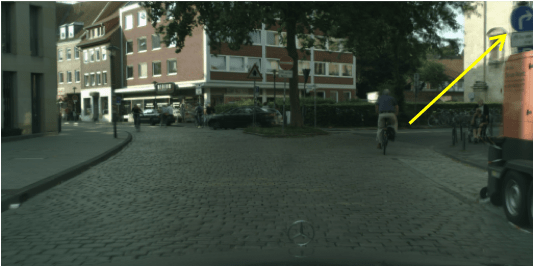}} &
  \multicolumn{2}{c}{\includegraphics[width=0.25\linewidth]{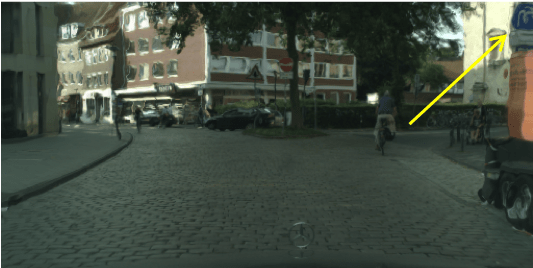}} &
  \multicolumn{2}{c}{\includegraphics[width=0.25\linewidth]{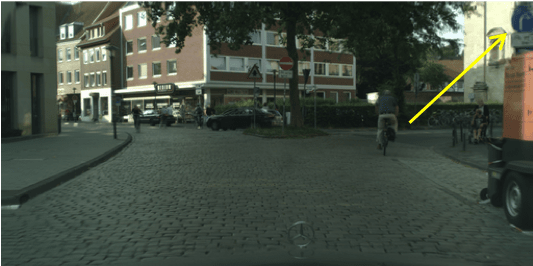}}
  \\
   \multicolumn{2}{c}{\includegraphics[width=0.25\linewidth]{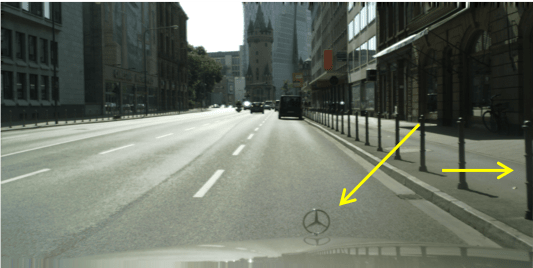}} &
  \multicolumn{2}{c}{\includegraphics[width=0.25\linewidth]{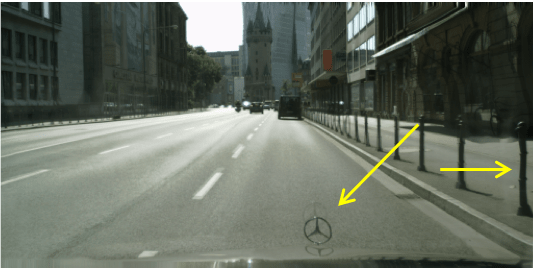}} &
  \multicolumn{2}{c}{\includegraphics[width=0.25\linewidth]{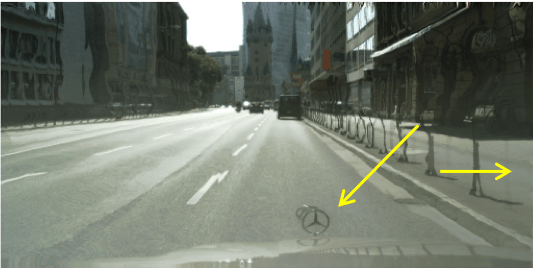}} &
  \multicolumn{2}{c}{\includegraphics[width=0.25\linewidth]{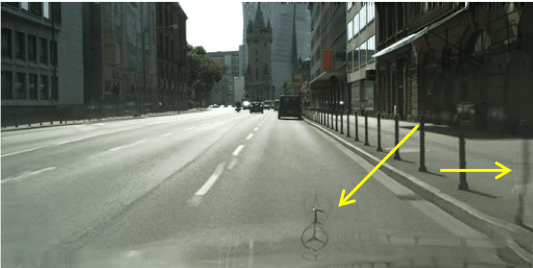}}
  \\
   \multicolumn{2}{c}{\includegraphics[width=0.25\linewidth]{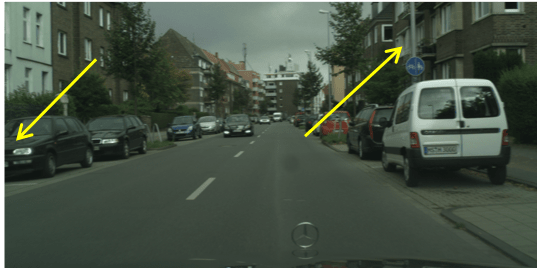}} &
  \multicolumn{2}{c}{\includegraphics[width=0.25\linewidth]{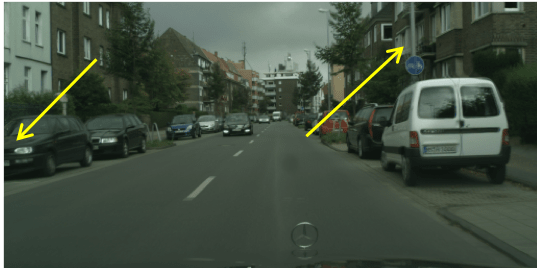}} &
  \multicolumn{2}{c}{\includegraphics[width=0.25\linewidth]{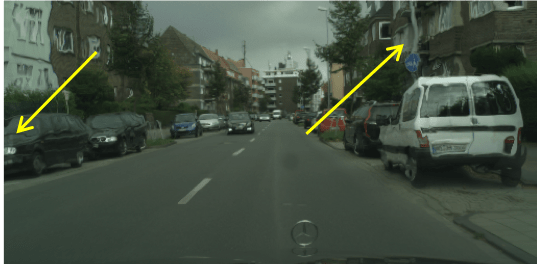}} &
  \multicolumn{2}{c}{\includegraphics[width=0.25\linewidth]{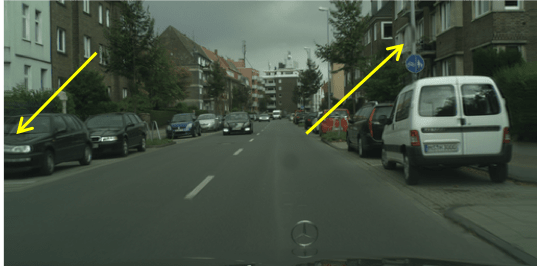}}
  \\
  \multicolumn{2}{c}{\includegraphics[width=0.25\linewidth]{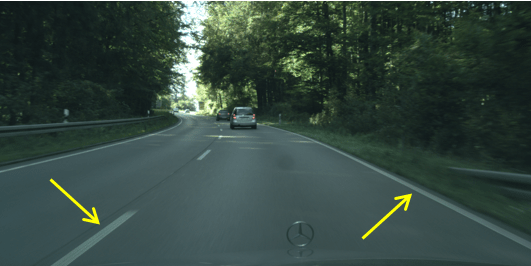}} &
  \multicolumn{2}{c}{\includegraphics[width=0.25\linewidth]{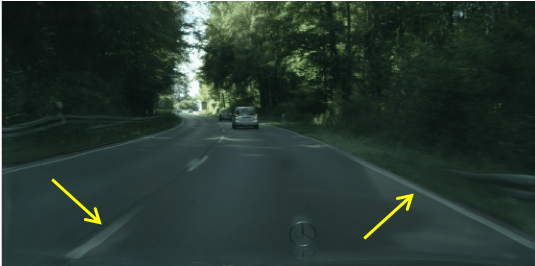}} &
  \multicolumn{2}{c}{\includegraphics[width=0.25\linewidth]{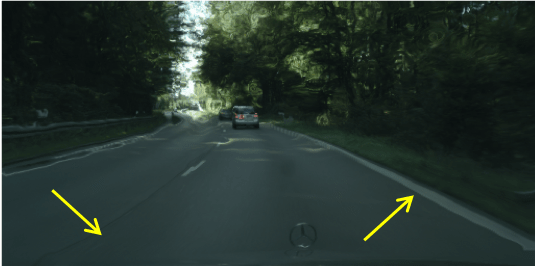}} &
  \multicolumn{2}{c}{\includegraphics[width=0.25\linewidth]{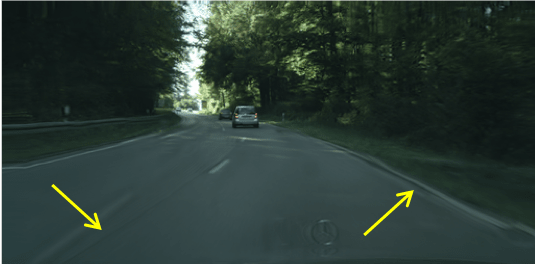}}
  \\
   \multicolumn{2}{c}{Ground Truth} &
   \multicolumn{2}{c}{Ours$_{roigan}$} &
   \multicolumn{2}{c}{DVF \cite{liu2017voxelflow}} &
   \multicolumn{2}{c}{SepConv \cite{Niklaus_ICCV_2017}}
  \end{tabular}
  \centering
  \caption{Qualitative results from different methods on CityScapes dataset. Best viewed in color.}
  \label{fig:compare_figs}
  \vspace{-15pt}
\end{figure*}

%% file: tex/exp/compare-ucf.tex
\begin{figure*}[t!]
\centering
\setlength\tabcolsep{1pt} % default value: 6pt
\begin{tabular}{cccccc}
 {\includegraphics[width=0.16\linewidth, height=0.16\linewidth]{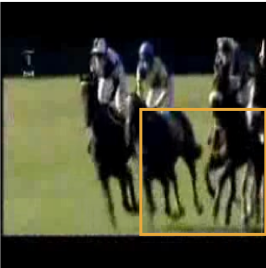}} &
 {\includegraphics[width=0.16\linewidth,height=0.16\linewidth]{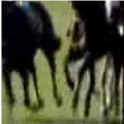}}&
 {\includegraphics[width=0.16\linewidth,height=0.16\linewidth]{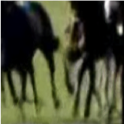}} &
 {\includegraphics[width=0.16\linewidth,height=0.16\linewidth]{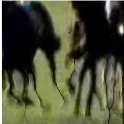}} &
  {\includegraphics[width=0.16\linewidth,height=0.16\linewidth]{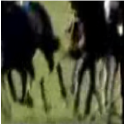}} &
  {\includegraphics[width=0.16\linewidth,height=0.16\linewidth]{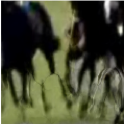}}
  \\
  {\includegraphics[width=0.16\linewidth,height=0.16\linewidth]{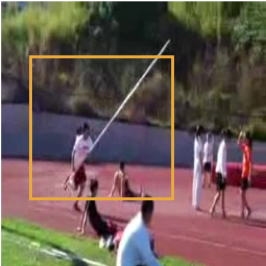}} &
  {\includegraphics[width=0.16\linewidth,height=0.16\linewidth]{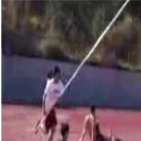}}&
  {\includegraphics[width=0.16\linewidth,height=0.16\linewidth]{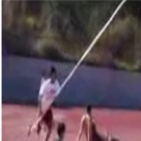}} &
  {\includegraphics[width=0.16\linewidth,height=0.16\linewidth]{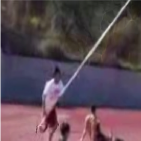}} &
  {\includegraphics[width=0.16\linewidth,height=0.16\linewidth]{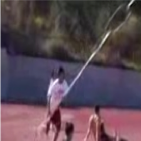}} &
  {\includegraphics[width=0.16\linewidth,height=0.16\linewidth]{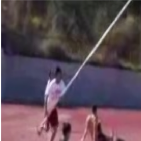}}
  \\
   {\includegraphics[width=0.16\linewidth,height=0.16\linewidth]{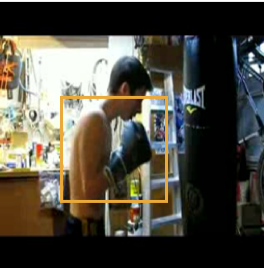}} &
   {\includegraphics[width=0.16\linewidth,height=0.16\linewidth]{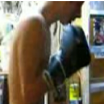}}&
  {\includegraphics[width=0.16\linewidth,height=0.16\linewidth]{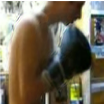}} &
  {\includegraphics[width=0.16\linewidth,height=0.16\linewidth]{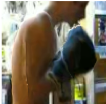}} &
  {\includegraphics[width=0.16\linewidth,height=0.16\linewidth]{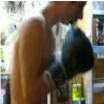}} &
  {\includegraphics[width=0.16\linewidth,height=0.16\linewidth]{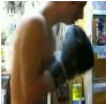}}
  \\
%   {\includegraphics[width=0.16\linewidth]{figs/compare-ucf/1/frame_01_gt_box.png}} &
%   {\includegraphics[width=0.16\linewidth]{figs/compare-ucf/1/gt_box.png}}&
%   {\includegraphics[width=0.16\linewidth]{figs/compare-ucf/1/frame_01_syth_box.png}} &
%   {\includegraphics[width=0.16\linewidth]{figs/compare-ucf/1/frame_01_dvf_box.png}} &
%   {\includegraphics[width=0.16\linewidth]{figs/compare-ucf/1/frame_01_sepconv_box.png}} &
%   {\includegraphics[width=0.16\linewidth]{figs/compare-ucf/1/frame_01_superslomo_box.png}}
%   \\
%   {\includegraphics[width=0.16\linewidth]{figs/compare-ucf/1531/frame_01_gt_box.png}} &
%   {\includegraphics[width=0.16\linewidth]{figs/compare-ucf/1531/gt_box.png}}&
%   {\includegraphics[width=0.16\linewidth]{figs/compare-ucf/1531/frame_01_syth_box.png}} &
%   {\includegraphics[width=0.16\linewidth]{figs/compare-ucf/1531/frame_01_dvf_box.png}} &
%   {\includegraphics[width=0.16\linewidth]{figs/compare-ucf/1531/frame_01_sepconv_box.png}} &
%   {\includegraphics[width=0.16\linewidth]{figs/compare-ucf/1531/frame_01_superslomo_box.png}}
%   \\
 {\includegraphics[width=0.16\linewidth,height=0.16\linewidth]{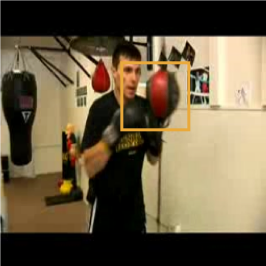}} &
 {\includegraphics[width=0.16\linewidth,height=0.16\linewidth]{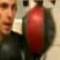}}&
  {\includegraphics[width=0.16\linewidth,height=0.16\linewidth]{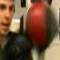}} &
  {\includegraphics[width=0.16\linewidth,height=0.16\linewidth]{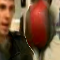}} &
  {\includegraphics[width=0.16\linewidth,height=0.16\linewidth]{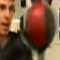}} &
  {\includegraphics[width=0.16\linewidth,height=0.16\linewidth]{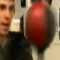}}
  \\
   {Ground Truth} &
   {GT Enlarged} &
   {Ours$_{roigan}$} &
   {DVF \cite{liu2017voxelflow}} &
   {SepConv \cite{Niklaus_ICCV_2017}} &
   {SuperSloMo \cite{jiang2017superslomo}}
  \end{tabular}
  \caption{Qualitative results from different methods on UCF101 dataset. Best viewed in color.}
  \label{fig:compare_ucf_figs}
  \vspace{-15pt}
\end{figure*}

%% file: tex/exp/Ablation.tex
\subsection{Ablation Studies}
\label{sec:ablation}
\textbf{Baseline.}
% In this section, we evaluate the effectiveness of different modules in our proposed system for video interpolation tasks and show both quantitative and qualitative results on CityScapes dataset. 
% We refer Ours$_{roigan}$ as our network trained with proposed instance-level adversarial loss, Ours$_{gan}$ as the model trained with the adversarial loss on overall image, and Ours$_{baseline}$ as the model trained without any adversarial loss.
% Note that without discriminator in the training stage, our model mostly degenerates to DVF \cite{liu2017voxelflow}.
% For all studies in this section, we trained the networks on mixed dataset consisting of cityscapes and UCF101 for 300k iterations and report the interpolation error (IE) \cite{Baker2011}, peak signal to noise ratio (PSNR) and structural similarity index (SSIM).
Considering our baseline is methodologically similar to \cite{ Niklaus_CVPR_2018,liu2017voxelflow}, comparing metrics within our methods (Ours$_{\text{baseline}}$, Ours$_{\text{roigan}}$) would serve as an ablation study to show the effectiveness of the proposed instance-level discrimination.
% coincides with the evaluation idea proposed by the reviewer. %so we would like to add one more discussion here.
% We measured all three metrics using the trimaps of \textit{`human', `vehicles'} groups with various dilation widths in CityScapes dataset. 
% Trimaps examples are shown in Fig.\ref{fig:example_trimap}, left.
% In Fig.\ref{fig:metrics_trimap}, we show our method achieves the best performance on object instances. When dilation width is less than 12 pixels, Ours$_{\text{roigan}}$ performs the best. As the trimap width grows to over 12 pixels, more background pixels are included such that Ours$_{\text{roigan}}$ performs slightly worse than SepConv \cite{Niklaus_ICCV_2017} on IE and PSNR. 
Results in both Tab.\ref{tab:merge} and Fig.\ref{fig:metrics_trimap} show Ours$_{\text{roigan}}$ consistently outperforms or on-par with Ours$_{\text{baseline}}$ across datasets. This shows from one aspect that the proposed instance-level discrimination can improve algorithm performance. 

% \textbf{Deformable convolution.} The deformable convolution layer in our image synthesis module is playing a substitute role for a well-designed but heavy network for synthesizing images. Practically, we can always find a large network to fine-tune the initial interpolation results warping by estimated optical flow and get a better result, which has been done in \cite{Niklaus_CVPR_2018}. However, utilizing a heavy network is wasting energy if we have a good flow estimation. It's an inefficient solution to this problem. Due to the flexible nature of deformable convolution, which explicitly learns offset map and warps features, we found a single layer deformable convolution is capable of learning complex deformation and correcting the incorrect region warped by flow estimation.

% ...
% further discussion.

\textbf{Adversarial Training.} 
We verify the advantages of using adversarial learning to improve video interpolation performance. 
From experiment on both datasets, training with adversarial loss gives us sharper boundaries in images. 
% In Table \ref{tab:cityscapes} and \ref{tab:ucf}, we find that our model with adversarial loss consistently outperforms the baseline model. 
In Fig.\ref{fig:zoom}, we show an example of the effectiveness of adversarial loss. 
From zoomed-in figures, we can see adversarial loss helps preserve edges and shapes. 
This can be attributed to the adversarial loss better facilitating image synthesis module's learning and potentially correcting the inaccurate optical flow estimation. 

Interestingly, we find training model with image level adversarial loss would lead to a local-minimal solution sometimes. This phenomenon is especially noted when testing on CityScapes dataset. %as shown in Fig.[], that network meets large uncertainty or complex deformation. 
Since the image-level adversarial training does not explicitly constrain the instances, the network tends to erase the uncertain objects in the scene and recover the background.
% , as shown in Fig.\ref{fig:failure}. 
This is because the data distribution is dominated by rigid objects and background such that training with image level adversarial loss leads to a biased learning result. 
In the next part, we discuss the proposed instance-level discrimination which would potentially fix this issue.

\textbf{Instance-level Discriminator.} 
We further verify the advantages of introducing an attention mechanism in adversarial training, which greatly improves the video interpolation performance as a result. 
From experiments, it is shown that training with instance-level discrimination gives us sharper boundaries on small, thin objects and image details.
With the adversarial loss, both rigid moving object and non-rigid human body shape are preserved better than baseline method with, as we can see in Fig.\ref{fig:zoom}.

In Table \ref{tab:merge}, we show Ours$_{roigan}$ method outperforms Ours$_{baseline}$ and Ours$_{gan}$ method on all three standards in CityScapes dataset.
% One typical failure case for applying adversarial losses to the whole image is that it is prone to erasing large non-rigid moving body parts as mentioned above.
In UCF101 dataset, as the RoI size in the image is quite close to the entire image size, the instance-level discriminator model and full-image-level discriminator model perform considerably similar.
Qualitatively, results in Fig.\ref{fig:compare_ucf_figs} still show better interpolation results on instances by our methods, due to the instance-level adversarial training.
Noticeably, we also measured all three metrics using the trimaps of \textit{`human', `vehicles'} groups with various dilation widths in CityScapes dataset to quantitatively illustrate the proposed instance-level discrimination improves synthesized instance quality.
Trimaps are generated using the groundtruth segmentation masks, as shown in Fig.\ref{fig:example_trimap}.
In Fig.\ref{fig:metrics_trimap}, we show our method achieves the best performance on object instances.
When dilation width is less than 12 pixels, Ours$_{\text{roigan}}$ performs the best.
As the trimap width grows to over 12 pixels, more background pixels are included such that Ours$_{\text{roigan}}$ performs slightly worse than SepConv \cite{Niklaus_ICCV_2017} on IE and PSNR. 
%This is because unlike in whole image GAN method, foreground bodies are only a small portion determining discriminator's decisions, in ROIGAN human bodies become major parts and force GAN discriminator to preserve body shape.
Introducing the region proposals and zooming into them force the network to focus on details and to utilize the fine-grained information for learning filters.
% instance-level semantics in the image. 
By formulating video interpolation problem as perturbing semantic objects in image space, the pixel-level motion estimation can be better grouped and updated.   

\textbf{Training with High Resolution Patches.} 
We also study the effects of training with different image resolution. 
Due to data augmentation and the concerns of training speed, researchers used to down-sample high-resolution images or crop part of images for training. 
However, high-resolution images often preserve fine-grained information and it can potentially improve algorithms performance. 
In our model, we train our proposed model with instance-level discriminator on real image patches from high-resolution images. 
More specifically, based on the region proposals we crop the `fake' RoIs from synthesized images and the corresponding `real' patches from its high-resolution counterpart, forming low-resolution high-resolution pairs.
The high-resolution patches ultimately force the generator to super-resolve and synthesize the details on low-resolution images.
From Table \ref{tab:merge}, we show that using high-resolution patches to train the network with region based adversarial training boosts performances beyond both the baseline model and the model using full image adversarial training. 
Fig.\ref{fig:metrics_trimap} also shows training with high-resolution image patches consistently improves interpolation qualities on instances.
%We only do this experiment on cityscapes dataset as images' resolution in UCF101 dataset is quite low.

%% file: tex/exp/Quantitative.tex
\subsection{Quantitative Evaluation}
\label{sec:quantitative}
We achieve the highest SSIM across datasets consistently, both on the foreground and full images.
We compare our approach with state-of-the-art video interpolation methods, including Separable adaptive Convolution (SepConv) \cite{Niklaus_ICCV_2017}, and Deep  Voxel Flow (DVF) \cite{liu2017voxelflow} on both UCF101 and CityScapes dataset.
As shown in Table \ref{tab:merge},
our method achieves the best SSIM score on CityScapes dataset. 
Table \ref{tab:merge} also demonstrates the quantitative results on UCF101 dataset, where we also compare with the SuperSloMo \cite{jiang2017superslomo}. We re-run the evaluation on the images provided by \cite{liu2017voxelflow} and \cite{jiang2017superslomo}, and images generated from \cite{Niklaus_ICCV_2017}. All metrics are computed under the motion masks provided by \cite{liu2017voxelflow}, which highlights the capabilities to cope with regions of motion and occlusion. Our method achieves the highest SSIM score among the lightweight models and performs comparably to the heavy model, SuperSloMo \cite{jiang2017superslomo}. 
We also show the proposed model achieves the highest SSIM on instances in Fig.\ref{fig:metrics_trimap}.
As the SSIM metric measures perceptual and structural similarity, it serves as a strong cue that the proposed method can render the most realistic scene and structural details as shown in Fig.\ref{fig:example_ssim}.

%% file: tex/exp/Qualitative.tex
\subsection{Qualitative Results}
\label{sec:qualitative}
We also present qualitative comparisons with other methods.
% to demonstrate the performance and advantages of our model. 
In Fig.\ref{fig:compare_figs}, %we show a comparison of our proposed approach with some state-of-the-art video frame interpolation methods. 
we present the comparison on different street scenes under various lighting condition. 
It is obvious that DVF\cite{liu2017voxelflow} generates the most artifacts such as distortion of the whole scene, unrealistic deformation of cars and buildings, misalignment of white lines and etc.. 
SepConv\cite{Niklaus_ICCV_2017} is capable of dealing with motion within their kernel size, but it consistently results in severe blur and artifacts near the image boundary, as shown in all of our examples. 
Our proposed approach is particularly good at recovering fine-grained details, for example, the traffic sign in the first example. 
Also, it fills up the occluded regions in a natural and realistic way, such as the white lines on the road in the fourth example. 
Fig.\ref{fig:compare_ucf_figs} shows a qualitative comparison on UCF101.
It is hard for DVF\cite{liu2017voxelflow} to handle the occlusion as shown in the second example, although it was trained on UCF101. 
SepConv\cite{Niklaus_ICCV_2017} is observed to have frequent duplicate artifacts, such as splits of horse legs and vault pole. 
SuperSloMo\cite{jiang2017superslomo} performs well in most scenes but sometimes fails in the refinement of details in small scale such as the chin of the boxing player, and legs of running horses.
Our proposed method enables the reconstruction of the fine-grained details and thus is capable of interpolating the challenging scenes.
% of fast motion, such as running horses and boxing. 

%% file: tex/exp/Discussion.tex
\subsection{Discussion}
\label{sec:disscusion}
\input{tex/exp/failure.tex}
Our network achieves the state-of-art video interpolation results using minimal model parameters and running fastest at inference time, illustrated in Fig.\ref{fig:runtime}. 
During training, we zoom into instances and train our model %with RoIs rescaled due to physical distance, which
by discriminating on the rescaled RoIs. 
The scaling on instances due to the physical distance and the zooming-in helps the network learn more structural and general filters that not only recover crispy boundary on objects but also structural patterns in the background, e.g. pole, traffic sign, etc. even though they are not explicitly trained by the discriminator.
Also, inspired by the super-resolution literature, we expect our model learn to super-resolve and render semantic details by training with high-resolution patches. 
% Besides instance-level adversarial loss, we employ other losses (Eq.5-7) on the full images to globally optimize the interpolation quality.
At inference time, only the flow estimation module and image synthesis module are needed, resulting in fast inference time.
% comparing to SuperSlomo\cite{jiang2017superslomo} using two-stage U-Nets, SDC-Net\cite{Reda2018sdcnet} using FlowNet2 and Context-aware Synthesis\cite{Niklaus_CVPR_2018} using an expensive GridNet. 
As a result, it costs our network 0.36s to run on a $1024 \times 2048$ image.
% , shown in Fig. \ref{fig:runtime}.
% while SDC-Net needs 1.66s and Context-aware Synthesis needs 0.77s to run on a $1920 \times 1080$ image. 
% In terms of visual comparison, besides keeping good interpolation results on rigid moving objects and background, our network works well on preserving sharp instance details like bicycle wheels, pen tips, etc. 
% The network recovers non-rigid body movement remarkably, keeping objects realistic when interpolating human faces, bodies, horse legs, etc..

Our network still has several limitations. For large non-rigid body movements, the interpolated objects are slightly distorted. 
As the right column in Fig.\ref{fig:failure} shows, cluttered scenes will lead to failure case. 
Large overlapping instances with mutual occlusion makes the system hard to dis-occlude individual objects. 
Adversarial learning is also likely to overfit to some data points in the training data. 
For example, when the motion estimation is blurry, the synthesis module tends to remove the uncertainty and choose to safely reconstruct the background, as shown in Fig.\ref{fig:failure} left column.
Finally, larger models with better optical flow estimation would generate better results than ours on a clean and texture-rich area (ground).

%% file: tex/exp/failure.tex
\begin{figure}[t!]
\centering
\setlength\tabcolsep{1pt} % default value: 6pt
\begin{tabular}{cc}
   {\includegraphics[width=0.49\linewidth]{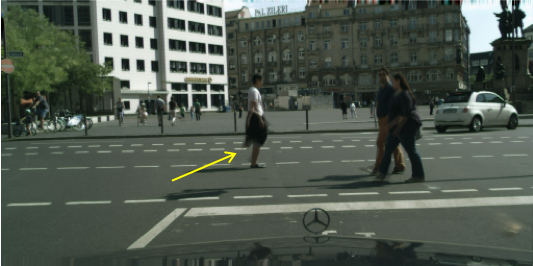}} &
  {\includegraphics[width=0.49\linewidth]{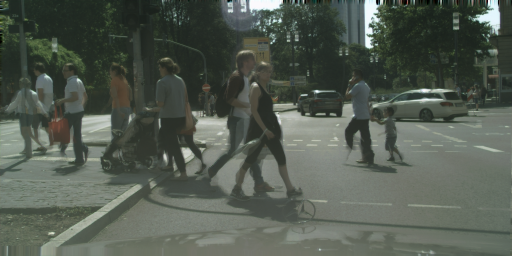}} \\
  [-0.5ex]
  {\includegraphics[width=0.49\linewidth]{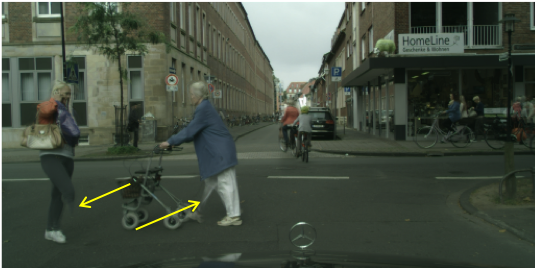}} &
  {\includegraphics[width=0.49\linewidth]{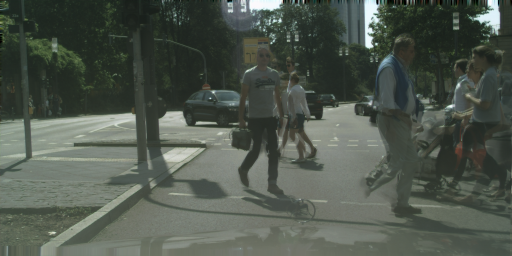}}
  \end{tabular}
  \caption{Failure cases on CityScapes dataset. Left Column: The network tends to erase objects and recover background to overfit to the training objectives. 
%   The man in the middle (upper-left), and the two ladies (lower-left) are left with one leg. 
  Right Column: Our model may fail in cluttered scenes.
%   due to fast movement and motion blur. 
%   (upper-right) The legs of the man in the right part of the image are cut while the legs of the son are fully reconstructed. 
%   (lower-right) One leg of a pedestrian is partially erased.
%   The head of the bicycle rider is erased to fully reconstruct the windshield of the van in the background. 
%   
%   Best viewed in color.
  }
  \label{fig:failure}
  \vspace{-15pt}
\end{figure}